\documentclass[conference]{IEEEtran}
\IEEEoverridecommandlockouts
\usepackage{cite}
\usepackage{amsmath,amssymb,amsfonts}
\usepackage{algorithm}
\usepackage{algorithmic}
\usepackage{graphicx}
\usepackage{textcomp}
\usepackage{xcolor}
\usepackage{xspace}
\usepackage{subfigure}
\usepackage{bbm}
\usepackage{hyperref}
\colorlet{kw}{blue}
\definecolor{com}{rgb}{0,0.6,0.3}

\newcommand{\netmask}{\emph{NeuroMask}\xspace}

\newcommand{\mm}{$\mathbf{m}$\xspace}

\usepackage{amsmath}
\DeclareMathOperator*{\argmax}{arg\,max}
\DeclareMathOperator*{\argmin}{arg\,min}

\def\BibTeX{{\rm B\kern-.05em{\sc i\kern-.025em b}\kern-.08em
    T\kern-.1667em\lower.7ex\hbox{E}\kern-.125emX}}
\begin{document}

\title{\emph{NeuroMask}: Explaining Predictions of Deep Neural Networks through Mask Learning
}

\author{\IEEEauthorblockN{ Moustafa Alzantot
\IEEEauthorblockA{\textit{Computer Science Dept.} \\
\textit{UCLA}\\
Los Angeles, USA \\
malzantot@ucla.edu}}
\and
\IEEEauthorblockN{
Amy Widdicombe 
\IEEEauthorblockA{\textit{Computer Science Dept.} \\
\textit{UCL}\\
London, UK \\
amy.widdicombe.17@ucl.ac.uk}}
\and
\IEEEauthorblockN{
Simon Julier
\IEEEauthorblockA{\textit{Computer Science Dept.} \\
\textit{UCL}\\
London, UK \\
s.julier@ucl.ac.uk }}
\and
\IEEEauthorblockN{Mani Srivastava
\IEEEauthorblockA{\textit{Computer Science Dept.} \\
\textit{UCLA}\\
Los Angeles, USA \\
mbs@ucla.edu}}
}

\maketitle
\IEEEpeerreviewmaketitle

\begin{abstract}
Deep Neural Networks (DNNs) deliver state-of-the-art performance in many image recognition and understanding applications. However, despite their outstanding performance, these models are black-boxes and it is hard to understand how they make their decisions. Over the past few years, researchers have studied the problem of providing explanations of why DNNs predicted their results. However, existing techniques are either obtrusive, requiring changes in model training, or suffer from low output quality. In this paper, we present a novel method, \netmask, for generating an interpretable explanation of classification model results. When applied to image classification models, \netmask identifies the image parts that are most important to classifier results by applying a mask that hides/reveals different parts of the image, before feeding it back into the model. The mask values are tuned by minimizing a properly designed cost function that preserves the classification result and encourages producing an interpretable mask. Experiments using state-of-art Convolutional Neural Networks for image recognition on different datasets (CIFAR-10 and ImageNet) show that \netmask successfully localizes the parts of the input image which are most relevant to the DNN decision. By showing a visual quality comparison between \netmask explanations and those of other methods, we find \netmask to be both accurate and interpretable.
\end{abstract}


\section{Introduction}
In the past decade, the world has witnessed a revolution in smart devices and machine intelligence. Computers in different forms and scales, ranging from servers to smart home devices, and from mobile phones to smart cars, are now achieving or exceeding human levels of autonomy and intelligence in certain specific situations. Much of this success has been made possible by the surge in the subset of machine learning algorithms known as Deep Neural Networks (DNNs). DNNs are a powerful way to learn approximations of the complex functions that underlie the process of decision making in many real-world applications (e.g. object recognition, image understanding, speech, and language understanding) that would have been hard for human and domain experts to tackle before. They also, most often, are trained using a domain-agnostic family of algorithms known as ``back propagation" and ``gradient descent" which require only using a labeled set of training examples and adjusting the model weight parameters to minimize an error, or \textit{loss} function, defined over the training examples and model weights. The final set of weights are used in the model to carry out future predictions.

While the existing algorithms for training and building DNNs work effectively and \textit{magically} to achieve unprecedented levels of accuracy in different application domains, they suffer from a major and, sometimes critical, limitation: DNNs lack the ability to provide explanations of how the predicted outcomes were computed. This is extremely important. Humans often have a prior knowledge as to what are the important types of evidence or features which support a particular decision. A machine learning algorithm, however, models the correlations between input features and classes. It can learn relationships which are purely the result of noise, biases in training data, or even poorly formed machine learning problems. For example, a cancer diagnosis model from MRI images should highlight which part of the MRI image looks abnormal. A job applicant selection program should be able to explain why it has preferred a given applicant over another. The explanation is needed to ensure the fairness and correctness of the deployed models and that they are not using spurious features. The need for this was dramatically illustrated in a recent incident where Amazon deployed an experimental artificial intelligence recruiting tool which rated every applicant on a scale from one to five based on their application materials \cite{amazonhiring}. Not too long after the experiment began, it was discovered that the program would systemically assign CVs of women applicants lower ratings than those of men with similar qualifications. The sexist failing behavior of the program is of course not intended and could have been spotted earlier if the model had been able to explain its results. For these reasons, governments around the world have started to create regulations (e.g. the General Data Protection Regulation (GDPR) ~\cite{GDPR}) requiring organizations that use machine learning and artificial intelligence to make decisions affecting users (such as approval of loans, hiring, etc.) to also provide an explanation of their decision. In a similar effort, the US government has organized the DARPA's explainable Artificial Intelligence (XAI)~\cite{gunning2017explainable} to encourage and promote research efforts that address this challenging problem. 

One method for understanding the performance of a learned algorithm is to look directly at its functional form. Some types of models, such as logistic regression and decisions trees, can sometimes be easy to understand. However, their results are often worse than those for DNNs. DNNs, on the other hand, learn models which are cascaded sequences of linear and non-linear operations (e.g. ReLU, tanh, sigmoid) which lead from the input to the final outcome. It is often impossible for a human to understand what evidence was used by the the model and how to reach its final conclusion. The limitation of DNNs to explain their outcomes and the evidence which supports them places barriers in their application in critical areas where an explanation is required and is as important as the outcome. This trade-off between model accuracy and interpretability urges the need for a robust approach to generate explanations of DNNs predictions.

\begin{figure*}[!th]
\centering
\includegraphics[clip, width=0.80\textwidth]{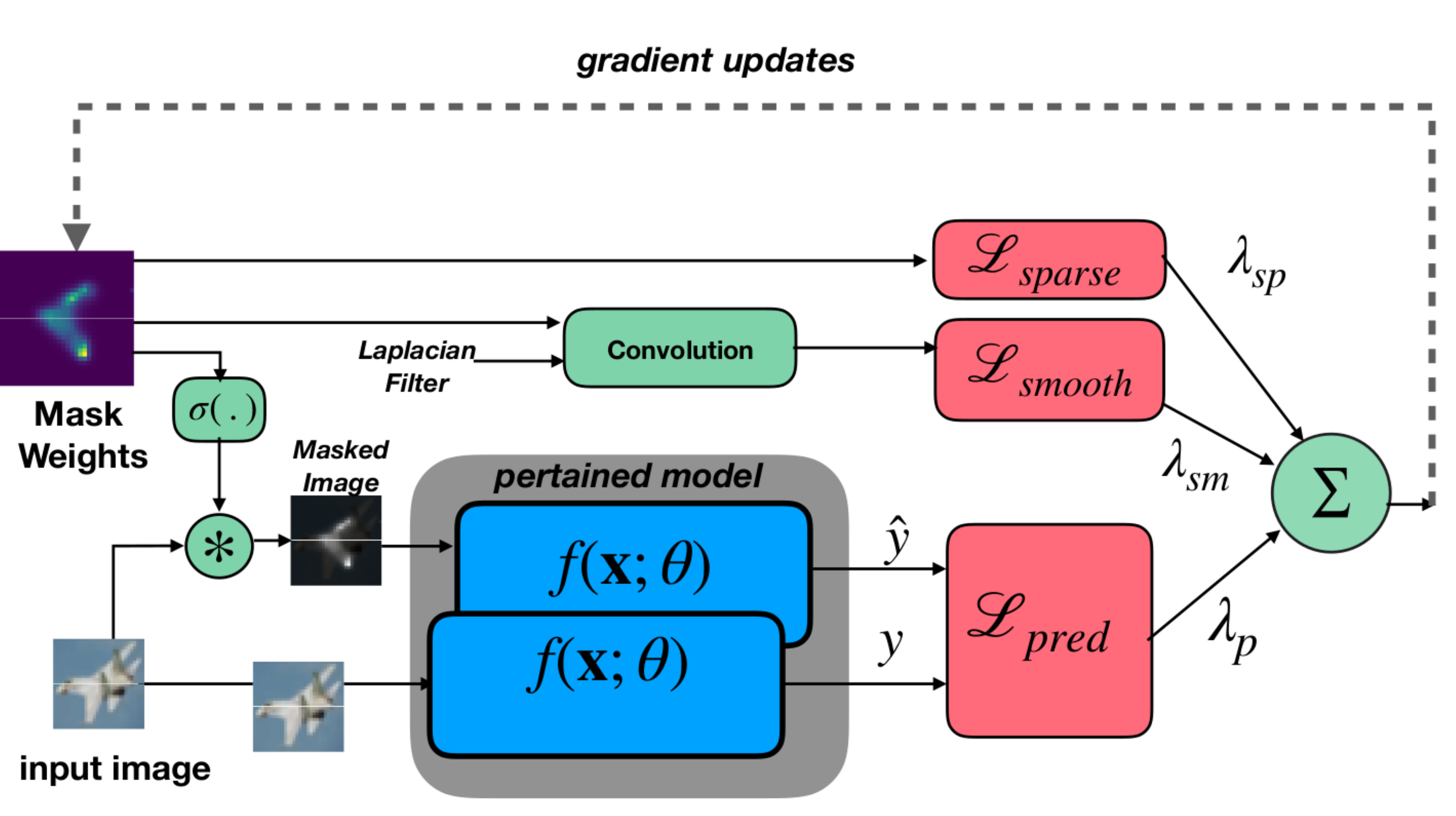}
\caption{Interactions of different components in \netmask. The blue blocks are clones of the pre-trained models. The dotted line represent gradient updates the adjust the values of mask weights.}
\label{fig:diagram}
\end{figure*}

Motivated by the above, many authors have developed intepretability methods based on heat maps: the parts of an input image area assigned weights to show their relative importance in a DNN decision. While these efforts have lead to the creation of a number of methods (e.g. Saliency Map~\cite{simonyan2013deep}, LIME~\cite{lime}, Smoothed Grad~\cite{smilkov2017smoothgrad}) as a way to explain neural networks results, they are still either computationally inefficient or generate noisy results. Our new method, \netmask, aims to generate better explanations by formulating the problem of generating explanations as an optimization problem to learn a `mask' that hides parts of the input which are irrelevant to the model output and leaves important parts still visible to the model. We introduce an efficient algorithm for learning this `mask' in Section~\ref{sec:alg}. In addition, when applied to images, we also add additional constraints that make the explanation results more interpretable for a human observer by promoting the mask to hide/reveal contiguous parts of the input image instead of individual pixels. 
\netmask does not require any modification to the architecture or training algorithm of the DNN model and can be applied to produce explanations for the outputs of any pre-trained model.

We demonstrate the effectiveness of \netmask by showing its explanations for the predictions of state-the-art ImageNet~\cite{deng2009imagenet} and CIFAR-10~\cite{krizhevsky2009learning} image recognition models on different examples. Visual comparison between \netmask and explanations generated by other methods (Saliency Map~\cite{simonyan2013deep}, Smoothed Grad~\cite{smilkov2017smoothgrad}, Guided Backprop~\cite{springenberg2014striving}, LIME~\cite{lime}, and LRP-epsilon~\cite{bach2015pixel}) reflect the success of \netmask to produce high-quality explanations.

The rest of this paper is organized as follows: Section~\ref{sec:alg} describes the assumptions and implementation of the \netmask algorithm. Section~\ref{sec:eval} describes our evaluation experiments and provides visual examples of explanations generated by \netmask. Section~\ref{sec:related} summarizes the related work. Finally, Section~\ref{sec:conc} concludes the paper and presents directions for our future work.


\section{Related Work}
\label{sec:related}
Over the past few years, researchers have studied the problem of DNNs interpretability.  While the definition of interpretability itself is still confusing as discussed in \cite{DBLP:journals/corr/Lipton16a}. Most of the work~\cite{smilkov2017smoothgrad, simonyan2013deep, zeiler2014visualizing, smilkov2017smoothgrad} use the term to refer how to explain the DNNs prediction results in terms of its own input which is the same definition we consider in our paper.

The major existing methods for interpretability can coarsely be categorized according to how they work into one of the following categories: \\
(a) \textbf{Occlusion-based}: such as ~\cite{zeiler2014visualizing}  which systemically occlude different parts of the input image with a grey square and monitor the change of the prediction class probability. This method while being effective is computationally inefficient due to its brute-force nature that requires trying occlusion square at every possible position in the input image. It is also unsuitable when objects in the image have different scales and arbitrary shapes. \\
(b) \textbf{Gradient-based}:  The Saliency Map~\cite{simonyan2013deep} method relies on computing the gradient of the prediction class label with respect to the model input to estimate features importance. However, the results of Saliency Map are often noisy and hard to interpret. To improve further,   The gradient-weighted action mapping (GRAD-CAM) ~\cite{selvaraju2017grad} uses the gradient of output label with respect to the final convolution layer to produce a coarse localization map highlighting important region in the image. Similarly, the SmoothGrad~\cite{smilkov2017smoothgrad} improves the quality of saliency maps by reducing the visual noise by using a regularization technique. The Layerwise Relevance Propagation (LRP) and Deep LIFT have been recently proposed as an alternative method. The difference between these two methods and prior work is that they attempt to estimate the global importance of pixels, rather than the local sensitivity.  LRP~\cite{10.1371/journal.pone.0130140}   was the first to propose the pixel-wise decomposition of classifiers in order to produce an explanation for a classification decision. They evaluate individual pixel contributions and produce ``interpretable'' heatmaps.  Guided-Backprop~\cite{springenberg2014striving} uses backpropagation and deconvolution operation to invert the computation of the DNN in order to visualize the concepts represents by intermediate layers. Guided-Backprop can be considered as equivalent to computing gradients except for the case when the gradient becomes negative then it will be zeroed out.\\
(c) \textbf{Approximate local model-based}: such as Local Interpretable Model-agnostic Explanations (LIME)~\cite{lime} algorithm. LIME~\cite{lime} generates an explanation of a model prediction of a given input by training another model (\textit{explainer}) which is selected from a group of intrinsically interpretable group of models (e.g. linear models, decision trees, etc.). The explainer is trained to approximate the model's decision surface around the provided input example. Training instances are obtained by drawing samples uniformly at random from the perturbed neighborhood of the input example. The perturbed samples are labeled with their prediction outcome from the given model and given the set of perturbed samples and their labels. The explainer is trained to mimic the decision surface of input model around the given example. LIME has been used to provide explanations for models of different data formats including text, tabular, and images. For images datasets, LIME uses super-pixels rather than individual pixels while producing the explainer training instances in order to produce a more interpretable explanation in terms of super-pixels. However, the reliance of superpixels sometimes can cause LIME to fail as we have observed in our experiments. It is also computationally inefficient due to the necessity of training the explainer model.
During our literature survey to prepare this manuscript, we have found the approach of~\cite{fong2017interpretable} to be the most similar to our idea. Both methods attempt to generate interpretable explanations of DNN decision by learning a mask that perturbs the input image. Nevertheless, our method is novel in its algorithm and cost function definition.

A more comprehensive review of literature in this topic can be found in~\cite{chakraborty2017interpretability} and ~\cite{guidotti2018survey}.

\section{Algorithm Design}
\label{sec:alg}

The basic idea behind \netmask is that input features which are not strongly relevant to the model's classification decision can be suppressed from the input without affecting the model's output. In order to find out which pixels are influential/unimportant, \netmask maintains two clones of the given pre-trained classification model. As an input to one of them, we feed the input example for which we seek an explanation of the model's result. The image fed into the other model copy is the input example after suppressing part of its features. By comparing the outputs of the two copies, \netmask attempts to find which are the unimportant features that could be suppressed while maintaining close similarity between the outputs of the two model copies. Importantly, the output explanation should be  \textit{comprehensible} for a human observer. Toward this end, we design \netmask such that the explanation output is both \textit{minimal} and \textit{interpretable}. In this rest of this section, we give more details on how \netmask operates.

We formalize the problem of explaining the predictions of a DNN from a given example, as follows:\\ Given
\begin{itemize}
    \item $\mathbf{x} \in \mathbb{R}^{H \times W \times 3}$: an input example representing RGB image whose height is $H$ and width is $W$.
    \item  $f(\mathbf{x}; \theta): \mathbb{R}^{H \times W \times 3} \rightarrow [0,1]^L $: a pre-trained image recognition model  that maps input example to one of different $L$ classification labels. $\theta$ denotes the set of model parameters.
\end{itemize} 
The goal of \netmask is to learn the values of a single channel mask filter $\mathbf{m} \in [0, 1]^{H \times W}$ with the same height and width as the input image. Elements of the mask are real values the $[0, 1]$ corresponding to the relevance of the input image pixels at the same coordinate to the model prediction. The value $1$ signifies a strongly relevant pixel and $0$ deemed as an irrelevant pixel.

Figure ~\ref{fig:diagram} illustrates the different components of \netmask and how they interact with each other. At the heart of \netmask, there are two copies of the given pre-trained model (shown within the grey box). The two copies are identical and their weights remain frozen during the operation of \netmask.  They only differ in the input applied to each one of them. Additionally, \netmask defines a set of trainable parameters which we refer to as `mask weights' $\mathbf{W}$. The mask weights are initialized from a uniform random distribution, and transformed into a `relevance mask' \mm using a \texttt{sigmoid} function. The relevance mask is what we need to compute by \netmask through the algorithm described in~\ref{alg:main}. One of the two model copies receives the input example as its input while the other receives the result of multiplying (\textit{pixel-wise}) the input example by the relevance mask. Outputs from the two models are fed into a cost function $\mathcal{L}_{pred}$ the measures the distance between their predictions. In addition, the mask weights $\mathbf{W}$ goes into  two additional cost terms $\mathcal{L}_{sparse}$ and $\mathcal{L}_{smooth}$ which are designed to improve the interpretability of the final mask weights. The weighted sum of these three terms ($\mathcal{L}_{pred}$, $\mathcal{L}_{sparse}$, $\mathcal{L}_{smooth}$) represent the total cost function $\mathcal{L}_{total}$ of \netmask.
\[
\mathcal{L}_{total} = \lambda_{p} \,\, \mathcal{L}_{pred} + \lambda_{sp} \,\,\mathcal{L}_{sparse} + \lambda_{sm} \,\, \mathcal{L}_{smooth} 
\]
where $\lambda_{p}$, $\lambda_{sp}$, and $\lambda_{sm}$ are weighting coefficient to balance between the different components of cost function.

We use the RMSProp~\cite{tieleman2012lecture} optimization algorithm to compute the final mask weights $\mathbf{W}^*$ that minimizes this cost function. Therefore, the final mask weights are defined by

\[ 
 \mathbf{W}^{*} = \argmin_{\mathbf{W}} \quad \mathcal{L}_{total}(\mathbf{W}; \mathbf{x}, \theta)
\]

The three components in \netmask cost function are defined according the the following:

\begin{itemize}
    \item \textbf{Prediction cost} $\mathcal{L}_{pred}$:  which measures the distance between the predictions of the two clones of the given model. It is defined as the cross-entropy between the two model copies outputs.
    \[ \mathcal{L}_{pred}(y, \hat{y}) = -\sum_{i \in \{1,2,..L\}}{\mathbbm{1}{(i = \argmax_{j \in \{1,2,..L\}}{ y_j})}\, \log{(\hat{y}_i)}}  \]
    
    \item \textbf{Sparseness cost} $\mathcal{L}_{sparse}$: A high-quality explanation should be minimal. Rather than declaring all pixels in the input image as relevant to the model's prediction, we need to identify as small as possible subset of pixels that are considered the most relevant. This can be achieved by forcing the ``relevance mask'' \mm to be sparse matrix. The sparseness of \mm can be encouraged by defining $\mathcal{L}_{sparse}$ as a $L_1$ regularizer over the mask weights.
    \[ \mathcal{L}_{sparse}(\mathbf{W}) = \sum_{i = 1}^{H}{\sum_{j=1}^{W}{ \vert \mathbf{W}_{i,j} + \tau \vert }}  \]
    This will force as many as possible elements of matrix $\mathbf{W}$ to be equal to $-\tau$. The $\tau$ (we pick its value  equal $20$) is chosen such that $\sigma(-\tau) \approx 0$.
    \item \textbf{Smoothness cost} $\mathcal{L}_{sparse}$:  In addition to being minimal, it is also desirable for an explanation to be \textit{interpretable}. I.e., the explanation would ideally be defined in terms of objects and object parts in the image rather than a subset of ungrouped pixels. Previous work (LIME, \cite{lime}) addressed this requirement by expressing the explanation in terms of super-pixels which are patches of nearby pixels with similar color intensities, the approach would often fail when we have nearby objects with similar colors (as we show in our results section). Inspired by the work of ~\cite{zhang2018interpretable} that altered the training algorithm of convolutional networks so that convolution filters correspond to interpretable parts of objects in the image, we employ a similar constraint that encourages the mask weights to be smooth and therefore highlights groups of spatially co-located pixels. Accordingly, the definition of $\mathcal{L}_{sparse}$ is chosen to be the $L_1$ norm of the 2\textsuperscript{nd} derivative of the mask weights, which are computed by convolving $\mathbf{W}$ with a discrete Laplacian filter $f_s$
    \begin{align*}
    \mathcal{L}_{smooth}(\mathbf{W}) = & {\vert  \nabla^2 \mathbf{W}(x, y) \vert}_1  = \left\lvert{ \frac{\delta^2 \mathbf{W}}{{\delta x}^2} + \frac{\delta^2 \mathbf{W}}{{\delta y}^2} }\right\vert_1,  \\
    = & \lvert \mathbf{W} \circledast f_s \rvert_1,
\end{align*}
where $\circledast$ denotes the 2d convolution operation.
\end{itemize}

Algorithm~\ref{alg:main} describes the steps to compute the explanation mask by minimizing the cost function.

\begin{algorithm}[!t]
   \caption{Optimization algorithm to compute the explanation mask $\mathbf{m}.$}
   \label{alg:main}
\begin{algorithmic}[1]
  \STATE {\bfseries Input:} input example $\mathbf{x} \in \mathbb{R}^{H \times W}$, a pre-trained prediction model $f: \mathbb{R}^d \rightarrow [0,1]^L$.
    \STATE {\bfseries Output:} $m$ explanation mask showing the relevance of different image parts to the model decision.
 \STATE $W \sim Uniform(-\tau, \tau)$  \COMMENT{Initialize mask weights}

\FOR{$i=1,...,T$ }
    \STATE $\mathbf{\bar{x}} = \sigma(\mathbf{W}) \cdot \mathbf{x}$ \COMMENT{Apply mask to input}
    \STATE $y = f(\mathbf{x}; \theta) $ \COMMENT{\text{Prediction of original input}}
    \STATE $\hat{y} = f(\mathbf{\bar{x}}; \theta ) $ \COMMENT{\text{Prediction of masked input}} 
    \STATE $\mathcal{L}_{pred} = -\sum_{i =1}^{L}{\mathbbm{1}{(i = \argmax_{j}{ y_j})}\, \log{(\hat{y}_i)}} $ 
    \STATE $\mathcal{L}_{sparse} = \lvert \mathbf{W} \rvert_1$
    \STATE $\mathcal{L}_{smooth} = \lvert \mathbf{W} \circledast f_s \rvert_1 $
    \STATE $d\mathbf{W} = \nabla_{\mathbf{W}} \left( \lambda_{p} \, \mathcal{L}_{pred} +   \lambda_{sp}\, \mathcal{L}_{sparse}  +  \lambda_{sm} \, \mathcal{L}_{smooth} \right) $
    \STATE \COMMENT {RMSProp optimizer step }
    \STATE $v_{dw} = \beta v_{dw}  + (1-\beta) {dW}^2$
    \STATE $W = W - \alpha \frac{dW }{\sqrt{dW}+\epsilon}$
\ENDFOR
\STATE $\mathbf{m} = \sigma(\mathbf{W})$ \COMMENT{Final mask}
\end{algorithmic}
\end{algorithm}

\section{Evaluation Results}
\label{sec:eval}

We evaluate the effectiveness of \netmask by demonstrating the visual quality of the explanations it generates of the predictions made by pre-trained state-of-art image recognition models. These models are tested using the CIFAR-10~\cite{krizhevsky2009learning} and ImageNet~\cite{deng2009imagenet} datasets. We also compare \netmask outputs to the outputs to other state-of-the-art interpretability methods.

\subsection{CIFAR-10 Results}

\begin{figure*}[!t]
\centering
\includegraphics[width=0.75\textwidth]{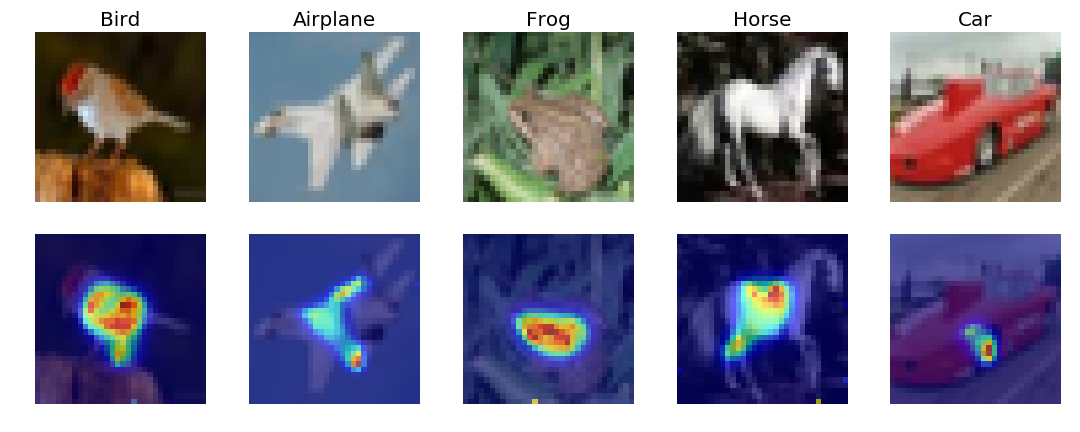}
\caption{Using \netmask to explain the predictions on a pre-trained model for image recognition on randomly selected examples from the CIFAR-10 dataset.}
\label{fig:cifar10}
\end{figure*}
The CIFAR-10 dataset~\cite{krizhevsky2009learning} contains small images (32x32 pixels) for 10 different categories. We used a convolutional network from ~\cite{carlini} that reaches near to state-of-the-art (80\%) classification accuracy on the CIFAR-10 dataset. We use \netmask to explain the classification model predictions on randomly selected images. As shown in figure~\ref{fig:cifar10}, \netmask can accurately localize the object within the image and highlight its discriminative parts (for example; tires of the car and wings of the airplane). This indicates that \netmask is effective in explaining the decisions of the pre-trained model.
\begin{figure*} [!t]
\centering
\includegraphics[width=0.90\textwidth]{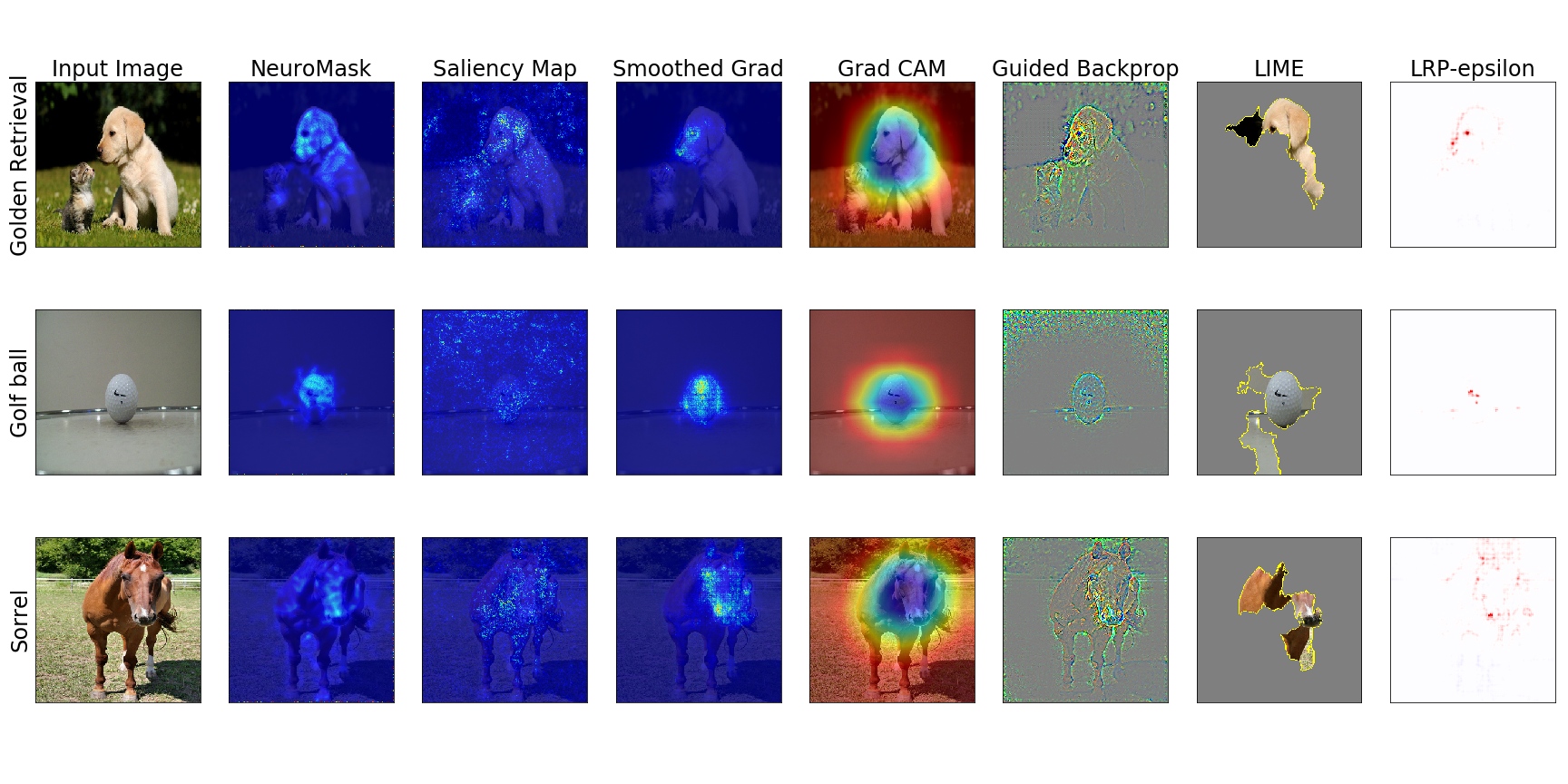}
\caption[]{Qualitative evaluation of the explanations make by \netmask vs other state-of-the-art interpretability methods (Saliency Map~\cite{simonyan2013deep}, Smoothed Grad~\cite{smilkov2017smoothgrad}, Grad-CAM~\cite{selvaraju2017grad}, Guided Backprop~\cite{springenberg2014striving}, LIME~\cite{lime}, LRP-epsilon~\cite{bach2015pixel}) on images selected from ImageNet~\cite{deng2009imagenet} test dataset using Inception\_v3 \cite{DBLP:journals/corr/SzegedyVISW15} image recognition model.}
\label{fig:Interp_Comparison}
\end{figure*}
\subsection{ImageNet Results}
We also performed comparison to evaluate the quality of explanations produced by \netmask versus prominent state-of-the-art methods (Saliency Map~\cite{simonyan2013deep}, Smoothed Grad~\cite{smilkov2017smoothgrad}, Grad-CAM~\cite{selvaraju2017grad}, Guided Backprop~\cite{springenberg2014striving}, LIME~\cite{lime}, and LRP-epsilon~\cite{bach2015pixel}) for DNNs interpretability. We randomly select test images from the ImageNet~\cite{deng2009imagenet} test dataset. The ImageNet dataset contains large scale ($299\times299$ pixels) images for 1000 different classes of images. In our experiments, We use the Inception-v3~\cite{MontavonBBSM15} as the pre-trained image recognition model (with 93.2\% top-5 accuracy). The implementation of LIME~\cite{lime} was obtained from the author's Github repo\footnote{\url{https://github.com/marcotcr/lime}}, while for the remainder of methods we use the implementations provided by the iNNvestigate tool kit \cite{DBLP:journals/corr/abs-1808-04260}.
 
 Our remarks from the visual comparison are as follow: both \netmask and Smoothed Grad~\cite{smilkov2017smoothgrad} produce explanations that are accurate and easy to interpret for a human observer. On the other hand, explanations produced by Saliency Map~\cite{simonyan2013deep} and Guided Backprop~\cite{springenberg2014striving} have too much noise which makes them hard to understand. By contrast, Grad-CAM~\cite{selvaraju2017grad} is too coarse grained while LRP-epsilon~\cite{bach2015pixel} is too conservative in what it highlights. Finally, LIME~\cite{lime} works well in some cases but fails in others (such as in the 2\textsuperscript{nd} row) where its reliance on superpixels causes it to declare the background as important as the golf-ball itself.
 
 Based on these results, we conclude that \netmask has a lot of potential and promise to provide explanations for DNNs outcomes. Nevertheless, in the future, we plan to conduct more comprehensive evaluation studies that include doing user's surveys to analyze their feedback on the different interpretability results.

\subsection{\netmask Learning Progress}
\begin{figure}[!h]
\centering
\includegraphics[width=0.85\columnwidth]{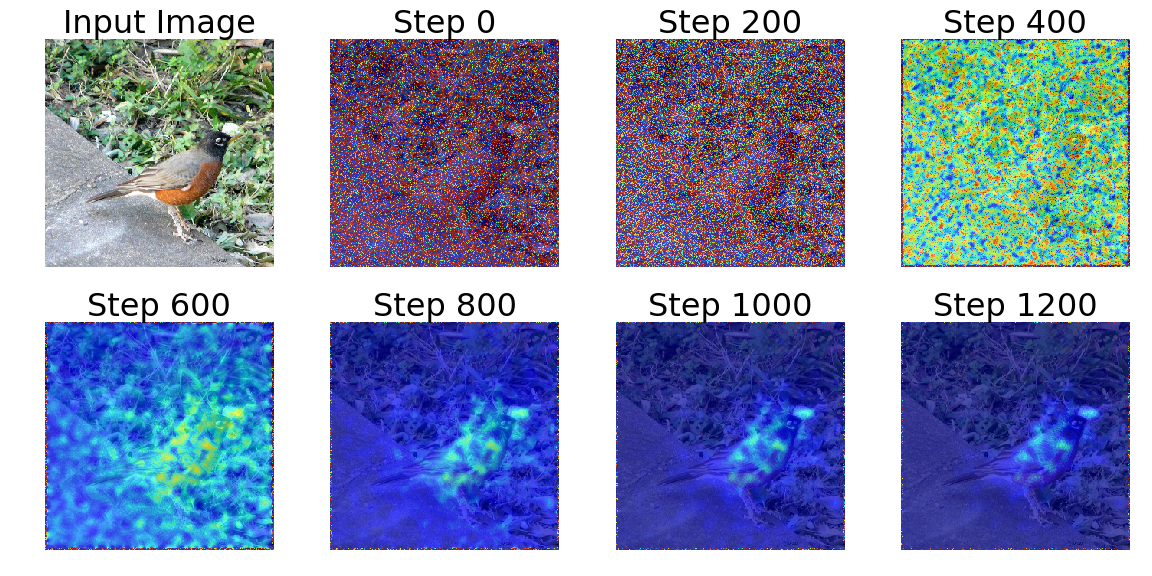}
\label{fig:progress}
\caption{Evolution of the explanation by \netmask during the optimization process.}
\end{figure}
Figure ~\ref{fig:progress} shows the progress of learning an interpretable explanation during the optimization process of \netmask. Since the mask weights are initialized as uniformly random, the mask initially looks like white noise (step 0) and then gradually it starts to focus more on the relevant part of the image as shown in Step 600. After that, the mask weights are refined to focus on important parts of the object which are most relevant to the classifier outcome.

\section{Conclusion}
\label{sec:conc}
In this paper, we presented \netmask a novel approach for generating explanations for the predictions of pre-trained deep neural networks. \netmask is model agnostic and can be used to explain the outputs of any image recognition models. The visual quality of explanations shows the success of \netmask to identify which parts of the input image were relevant to the classifier decision. Compared to explanations generated by other interpretability methods, we find \netmask to produce competitive explanations. Our directions for future work include: extending \netmask to DNNs used for other tasks such as image captioning, and data modalities such as text and sound.  We are also going to conduct users study surveys to understand their perception and feedback of the \netmask generated explanations.

\section*{Acknowledgment}
This research was supported in part by  the U.S. Army Research Laboratory and the UK Ministry of Defence under Agreement Number W911NF-16-3-0001, the National Science Foundation under award \# CNS-1705135. Any findings in this material are those of the author(s) and do not reflect the views of any of the above funding agencies. The U.S. and U.K. Governments are authorized to reproduce and distribute reprints for Government purposes notwithstanding any copyright notation hereon.

\bibliographystyle{IEEEtran}
\bibliography{IEEEabrv,references}

\end{document}